\documentclass[10pt,twocolumn,letterpaper]{article}

\usepackage{cvpr}              

\usepackage{graphicx}
\usepackage{amsmath}
\usepackage{amssymb}
\usepackage{amsbsy}
\usepackage[accsupp]{axessibility}
\usepackage{bbding}
\usepackage{bm}
\usepackage{booktabs}
\usepackage{caption}
\usepackage{multirow}
\usepackage{makecell}
\usepackage{nccmath}

\usepackage[pagebackref,breaklinks,colorlinks]{hyperref}

\usepackage[capitalize]{cleveref}
\crefname{section}{Sec.}{Secs.}
\Crefname{section}{Section}{Sections}
\Crefname{table}{Table}{Tables}
\crefname{table}{Tab.}{Tabs.}
\Crefname{equation}{Equation}{Equations}
\crefname{equation}{Eqn.}{Eqns.}

\def\cvprPaperID{4801} 
\def\confName{CVPR}
\def\confYear{2023}

\begin{document}


\title{Spatially Adaptive Self-Supervised Learning for Real-World Image Denoising}

\author{Junyi Li$^1$, Zhilu Zhang$^1$, Xiaoyu Liu$^1$, Chaoyu Feng, Xiaotao Wang, Lei Lei, Wangmeng Zuo$^{1,2(}$\Envelope$^)$\\
$^1$School of Computer Science and Technology, Harbin Institute of Technology, China\\
$^2$Peng Cheng Laboratory, China\\
{\tt\small nagejacob@gmail.com, cszlzhang@outlook.com, liuxiaoyu1104@gmail.com, wmzuo@hit.edu.cn}
}
\maketitle

\begin{abstract}
Significant progress has been made in self-supervised image denoising (SSID) in the recent few years.
However, most methods focus on dealing with spatially independent noise, and they have little practicality on real-world sRGB images with spatially correlated noise.
Although pixel-shuffle downsampling has been suggested for breaking the noise correlation, it breaks the original information of images, which limits the denoising performance.
In this paper, we propose a novel perspective to solve this problem, \ie, seeking for spatially adaptive supervision for real-world sRGB image denoising.
Specifically, we take into account the respective characteristics of flat and textured regions in noisy images, and construct supervisions for them separately.
For flat areas, the supervision can be safely derived from non-adjacent pixels, which are much far from the current pixel for excluding the influence of the noise-correlated ones.
And we extend the blind-spot network to a blind-neighborhood network (BNN) for providing supervision on flat areas.
For textured regions, the supervision has to be closely related to the content of adjacent pixels.
And we present a locally aware network (LAN) to meet the requirement, while LAN itself is selectively supervised with the output of BNN.
Combining these two supervisions, a denoising network (\eg, U-Net) can be well-trained.
Extensive experiments show that our method performs favorably against state-of-the-art SSID methods on real-world sRGB photographs.
The code is available at \url{https://github.com/nagejacob/SpatiallyAdaptiveSSID}.
\end{abstract}

\section{Introduction}
\begin{figure}[t]
    \centering
    \begin{subfigure}[b]{0.49\linewidth}
        \centering
        \includegraphics[width=\linewidth]{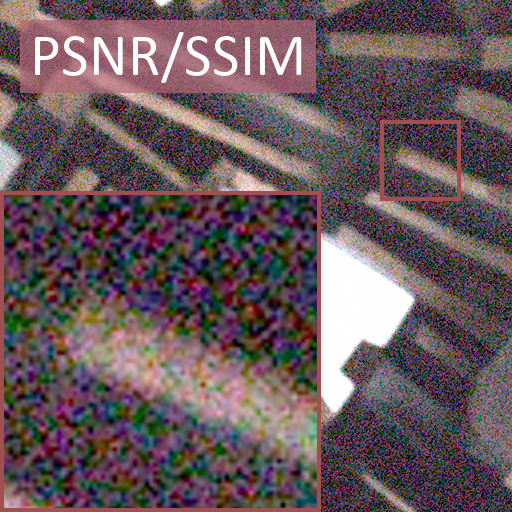}
        \caption{\centering\footnotesize Noisy Input}
    \end{subfigure}
    \hfill
    \begin{subfigure}[b]{0.49\linewidth}
        \centering
        \includegraphics[width=\linewidth]{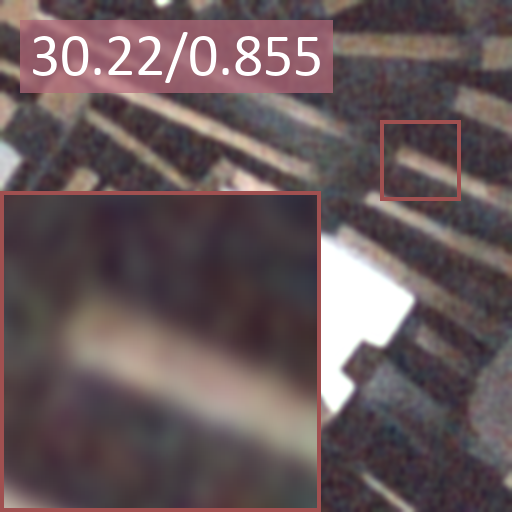}
        \caption{\centering\footnotesize CVF-SID~\cite{neshatavar2022cvf}}
    \end{subfigure}
    \hfill
    \begin{subfigure}[b]{0.49\linewidth}
        \centering
        \includegraphics[width=\linewidth]{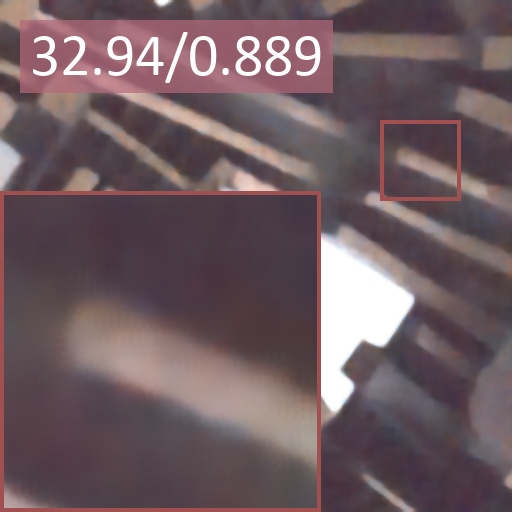}
        \caption{\centering\footnotesize AP-BSN+R$^3$~\cite{lee2022ap}}
    \end{subfigure}
    \hfill
    \begin{subfigure}[b]{0.49\linewidth}
        \centering
        \includegraphics[width=\linewidth]{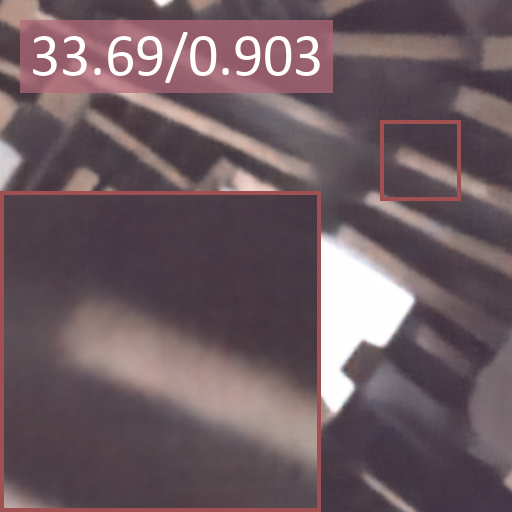}
        \caption{\centering\footnotesize Ours}
    \end{subfigure}
    \caption{Visual comparison between self-supervised denoising methods on the DND dataset~\cite{plotz2017benchmarking}. PSNR (dB) and SSIM with respect to the ground-truth are marked on the result for quantitative comparison. Our method performs better in removing spatially correlated noise from real-world sRGB photographs.}
    \label{fig:intro-dnd}
\end{figure}
Image denoising aims to restore clean images from noisy observations\cite{buades2005non, dabov2007image, gu2014weighted}, and it has achieved noticeable improvement with the advances in deep networks~\cite{zhang2017beyond, zhang2018ffdnet, mao2016image, tai2017memnet, liu2018non, liu2018multi, anwar2019real, yue2019variational, kim2020transfer, yue2020dual, ren2021adaptive, cheng2021nbnet, liang2021swinir, zamir2022restormer, wang2022uformer}.
However, the models trained with synthetic noise usually perform poorly in real-world scenarios in which noise is complex and changeable. 
A feasible solution is to collect real-world clean-noisy image pairs~\cite{plotz2017benchmarking, abdelhamed2018sidd} and take them for model training~\cite{guo2019toward, anwar2019real, kim2020transfer, yue2019variational, yue2020dual}.
But building such datasets generally requires strictly controlled environment as well as complicated photographing and post-processing, which is time-consuming and labor-intensive.
Moreover, the noise statistics vary under different cameras and illuminating conditions~\cite{wei2020physics, zhang2021rethinking}, and it is impractical to capture pairs for every device and scenario.

To circumvent the limitations of noisy-clean pairs collection, self-supervised image denoising (SSID) approaches~\cite{lehtinen2018noise2noise, krull2019noise2void, batson2019noise2self, laine2019high, moran2020noisier2noise, xu2020noisy, pang2021recorrupted, kim2021noise2score, kim2022noise, huang2021neighbor2neighbor, wang2022blind2unblind, lee2022ap, neshatavar2022cvf} have been proposed, which can be trained merely on noisy images.
However, the noise model assumptions of a large amount of SSID methods do not match the characteristics of real-world noise in sRGB space.
For instance, HQ-SSL~\cite{laine2019high} improves the denoising performance with posterior inference, but requires explicit noise probability density.
Noise2Score~\cite{kim2021noise2score} and its extension~\cite{kim2022noise} propose a closed-form image denoising schema with score matching followed by noise model and noise level estimation, but the noise is bounded to Tweedie distribution.
Although some methods~\cite{huang2021neighbor2neighbor, wang2022blind2unblind} are designed for distribution agnostic noise, they can only deal with spatially independent noise.

Recently, a few attempts have been explored to remove spatially correlated noise in a self-supervised manner.
CVF-SID~\cite{neshatavar2022cvf} disentangles the image and noise components from noisy images, but the difficulty of optimization limits its performance.
Some methods~\cite{zhou2020awgn, wu2020unpaired} break the spatial noise correlation with pixel-shuffle downsampling (PD), then utilize spatially independent denoisers (\eg, blind-spot network~\cite{laine2019high, wu2020unpaired, byun2021fbi}) to remove the uncorrelated noise.
However, PD breaks the original information of the images and leads to aliasing artifacts, which largely degrade the image quality.
AP-BSN~\cite{lee2022ap} applies asymmetric PD factors and post-refinement processing to seek for a better trade-off between noise removal and aliasing artifacts, but it is time-consuming during inference.

In this paper, we present a novel perspective for SSID by considering the respective characteristics of flat and textured regions in noisy images, resulting in a spatially adaptive SSID method for real-world sRGB images.
Instead of utilizing pixel-shuffle downsampling and blind-spot network to learn denoising results directly, we seek for spatially adaptive supervision for a denoising network (\eg, U-Net~\cite{ronneberger2015u}).
Concretely, for flat areas, the supervision can be safely derived from non-adjacent pixels, which are much far from the current pixel for excluding the influence of noise correlation.
We achieve it by extending the blind-spot network (BSN)~\cite{laine2019high} to a blind-neighborhood network (BNN). 
BNN modifies the architecture of BSN to expand the size of blind region, and takes the same self-supervised training schema as BSN.
%
%
%
%
Note that it is difficult to determine whether an area is flat or not from the noisy images, so we directly apply BNN to the whole image and it has little effect on the handling of flat areas.
Moreover, such an operation can give us a chance to detect textured areas from the output of BNN, whose variance is usually higher.
For textured areas, neighboring pixels are essential for predicting the details and they can not be ignored.
To this end, we present a locally aware network (LAN), which focuses on recovering the texture details solely from adjacent pixels.
LAN is supervised by flat areas of BNN output.
When training is done, LAN will be applied to textured areas to generate supervision information for these areas.
Combining the learned supervisions for flat and textured areas, a denoising network can be readily trained.
During inference, BNN and LAN can be detached,  only the ultimate denoising network is used to restore clean images.
Extensive experiments are conducted on SIDD~\cite{abdelhamed2018sidd} and DND~\cite{plotz2017benchmarking} datasets.
The results demonstrate our method is not only effective but also efficient.
In comparison to state-of-the-art self-supervised denoising methods, our method behaves favorably in terms of both quantitative metrics and perceptual quality.
The contributions of this paper can be summarized as follows:
\begin{itemize}
    \item We propose a novel perspective for self-supervised real-world image denoising, \ie, learning spatially adaptive supervision for a denoising network according to the image characteristics. 
    \item For flat areas, we extend the blind-spot network to a blind-neighborhood network (BNN) for providing supervision information. For texture areas, we present a locally aware network (LAN) to learn that from neighboring pixels.
    \item Extensive experiments show our method has superior performance and inference efficiency against state-of-the-art SSID methods on real-world sRGB noise removal.
\end{itemize}

\section{Related Work}
%
%
\subsection{Deep Image Denoising}
The development of convolutional neural networks (CNNs) has led to great improvement on deep image denoising~\cite{zhang2017beyond, zhang2018ffdnet}.
DnCNN~\cite{zhang2017beyond} outperforms traditional patch-based methods~\cite{buades2005non, dabov2007image, gu2014weighted} on Gaussian denoising.
FFDNet~\cite{zhang2018ffdnet} takes a noise level map as input, which can handle various noise levels with a single model.
Other learning based methods, such as RED30~\cite{mao2016image}, MemNet~\cite{tai2017memnet}, and MWCNN~\cite{liu2018multi}, are also developed with advanced architectures.
Nevertheless, models trained on AWGN generalize poorly to real scenarios due to the domain discrepancy between synthetic and real noise.

In order to mitigate the gap between synthetic and real noise, one feasible way is to simulate realistic noise as much as possible and introduce it during the network training~\cite{guo2019toward, zamir2020cycleisp}.
For example, CBDNet~\cite{guo2019toward} inverses the demosaicing and gamma correction steps in image signal processing (ISP), then synthesizes signal-dependent Poisson-Gaussian noise~\cite{foi2008practical} in  raw space.
Zhou~\textit{et~al.}~\cite{zhou2020awgn} break spatial-correlated noise into pixel-independent one with pixel-shuffle downsampling, then handle it with AWGN-based denoiser.
Another way is to capture paired noisy-clean images for constructing real-world datasets~\cite{plotz2017benchmarking, abdelhamed2018sidd}.
Taking such datasets for training, models have better potential to generalize to the corresponding real noise~\cite{anwar2019real, yue2019variational, kim2020transfer, yue2020dual, ren2021adaptive, cheng2021nbnet, liang2021swinir, zamir2022restormer, wang2022uformer}.
However, conducting well-aligned training pairs generally requires a controlled environment and much human labor.
Moreover, noise statistics vary under different cameras and different illuminating conditions~\cite{wei2020physics, zhang2021rethinking}. 
It is impractical to collect datasets for every device and scenario.

\subsection{Unpaired Image Denoising}
%
%
Unpaired methods~\cite{chen2018image, hong2020end, wu2020unpaired, jang2021c2n} mitigate the data collecting issue by training networks with unpaired noisy and clean datasets.
Most of which~\cite{chen2018image, hong2020end, jang2021c2n} learn to model the noise statistics with adversarial training.
A noise generator is first trained to match the noise distribution of the noisy images, then used to map the clean images to pseudo-noisy ones.
Finally, the denoising network is trained with synthetic pseudo-noisy and clean image pairs.
In addition, Wu~\textit{et~al.}~\cite{wu2020unpaired} learn the noise distribution by jointly training a denoising network and a noise estimator.
And the ultimate denoising network is trained with pseudo-noisy and clean pairs as well as noisy and denoised pairs.
However, as the noise distribution in sRGB space is very complex and difficult to model~\cite{kousha2022modeling}, the performance of unpaired methods is still limited when facing real-world photographs.

\subsection{Self-Supervised Image Denoising}
%
In order to get rid of dependence on clean images, self-supervised methods are proposed, which are trained with noisy images only.
Noise2Noise~\cite{lehtinen2018noise2noise} suggests to learn a model from paired noisy images, which remains limited in practice.
Subsequently, Noise2Void~\cite{krull2019noise2void} and Noise2Self~\cite{batson2019noise2self} separate noisy images into input and target pairs with a mask strategy.
Blind-spot networks take a step further by excluding the corresponding input noisy pixel from the receptive field for each output pixel, which can be implemented with multiple network branches~\cite{laine2019high, cha2019fully} or dilated and masked convolutions~\cite{wu2020unpaired, byun2021fbi}.
Moreover, probabilistic inference~\cite{laine2019high, krull2020probabilistic} and regular loss functions~\cite{wang2022blind2unblind} are proposed to alleviate the information loss issue at the blind spot.
In addition, Noisier2Noise~\cite{moran2020noisier2noise} and NAC~\cite{xu2020noisy} take the noisy images as target, and synthesize and add new noise to the noisy images for model training.
More recently, Noise2Score~\cite{kim2021noise2score} and its extension~\cite{kim2022noise} propose a closed-form denoising for Tweedie distributions with score matching and posterior inference.
SelfIR~\cite{SelfIR} introduce blurry images for self-supervised denoising task.
Nevertheless, the above self-supervised methods can only handle noisy images with spatially independent noise.

A few attempts have been done to remove the spatially correlated noise in a self-supervised manner.
R2R~\cite{pang2021recorrupted} reverses the noisy images to raw space to synthesize training pairs, then renders the synthetic raw images back to sRGB space.
But it requires several priors, such as the parameters of the camera ISP and noise model.
StructN2V~\cite{broaddus2020removing} adapts the blind mask from a single pixel~\cite{krull2019noise2void} to match the structure of the noise.
CVF-SID~\cite{neshatavar2022cvf} disentangles noisy images into clean images and noise components.
Among the self-supervised approaches for real-world sRGB noise, AP-BSN~\cite{lee2022ap} propose asymmetric PD factors and post-refinement processing to make a better trade-off between noise removal and aliasing artifacts, but it is time-consuming during inference.

\section{Method}
Self-supervised denoising aims to predict the denoised images $\hat{\textbf{x}}$ from the noisy observations $\textbf{y}$, without the supervision from clean images $\textbf{x}$.
In this paper, we propose a new perspective for self-supervised denoising.
Specifically, we extract appropriate supervisions from noisy images according to the image characters (\ie, flatness) for denoising network training. 
%
%
In this section, we first introduce how to learn the supervision for flat (Sec.~\ref{sec:flat_sup}) and textured (Sec.~\ref{sec:text_sup}) areas, respectively.
Then we discuss how to utilize the supervisions for the denoising network in Sec.~\ref{sec:denoising_network_training}.

\begin{figure*}[t!]
    \centering
    \includegraphics[width=0.95\linewidth]{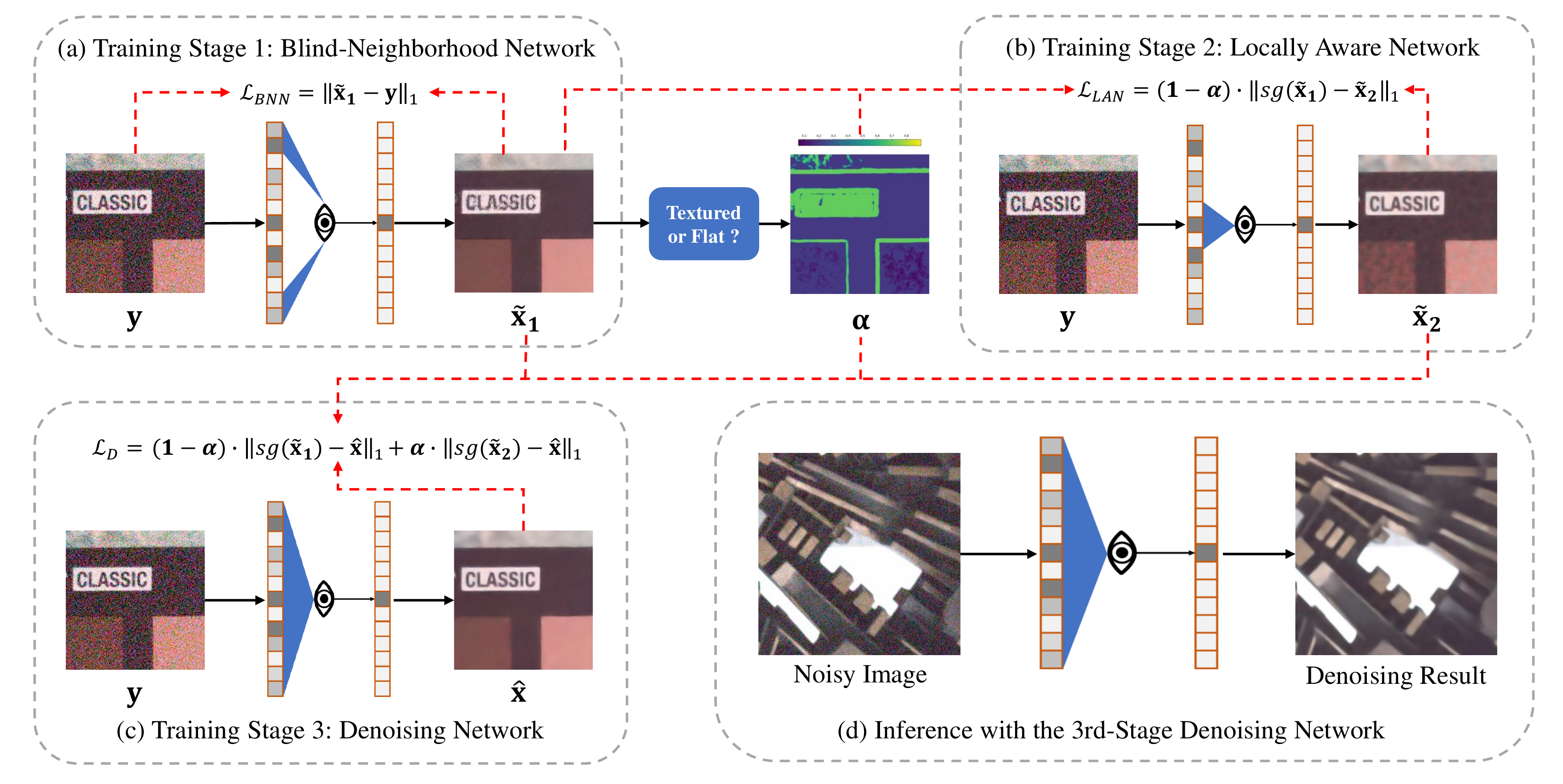}
    \caption{Overview of our self-supervised denoising framework. (a) In training stage one, the blind-neighborhood network (BNN) learns to remove the spatially correlated noise for flat areas in a self-supervised manner. (b) In training stage two, the locally aware network (LAN) is supervised by flat areas of BAN outputs. When training is done, LAN will be applied to texture areas to generate texture details. (c) In training stage three, the denoising network is supervised by the first two stages' results with adaptive coefficients. (d) During inference, the 3rd-stage denoising network can be deployed to denoise real-world photographs.}
    \label{fig:overview}
\end{figure*}

\subsection{Supervision for Flat Areas} \label{sec:flat_sup}
For a noisy image, denoising of flat areas is generally easier than that of textured areas whose edges and details need to be preserved well~\cite{zhao2019texture}.
And the same goes for our extraction of supervision information.
Thus, we handle these two areas separately with the consideration of their different characteristics.
In this subsection, we first learn the supervision for flat areas.

For removing spatially correlated noise in real-world sRGB images, pixel-shuffle downsampling (PD) has been applied to break the noise correlation~\cite{zhou2020awgn, wu2020unpaired, lee2022ap}.
Then blind-spot network (BSN)~\cite{laine2019high, wu2020unpaired, byun2021fbi} can be applied to the downsampled sub-images to remove the spatially independent noise.
However, the high-frequency information is distorted during downsampling~\cite{gonzalez2009digital}, leading to aliasing artifacts.
Moreover, apart from noise-correlated neighbor pixels, a large amount of non-neighboring pixels are removed, which degrades the denoising performance.

In terms of flat areas in noisy images, a pixel of the underlying clean images is similar to a large number of surrounding pixels, not only the neighboring pixels but also non-neighboring ones.
The non-neighboring pixels can be far enough from the current pixel that exceeds the noise correlation range.
Thus, instead of PD, we can enlarge the blind-spot of BSN into a blind-neighborhood until covering all noise-correlated pixels.
We present a blind-neighborhood network (BNN) to achieve it, as shown in Figure~\ref{fig:overview}(a).
Different from PD, BNN can make full use of the uncorrelated noisy pixels and preserve the original information as much as possible.
Similar to BSN, the loss function of BNN is as follows:
\begin{equation}
    \mathcal{L}_{BNN} = \left\| \tilde{\textbf{x}}_1 - \textbf{y} \right\|_1,
\end{equation}
where $\hat{\textbf{x}}_1$ is the output of BNN, $\textbf{y}$ is the noisy image.
As the random noise can not be predicted, the output of BNN will converge to the clean image.

In addition, the network architecture of BNN is modified from the BSN used in HQ-SSL~\cite{laine2019high}.
The BSN applies four network branches whose receptive fields are restricted in different directions.
At the end of each branch, a single pixel shift is applied to the features to create the blind-spot.
We increase the pixel shift size from 1 to $k$ to create a $(2k\!-\!1)\!\times\!(2k\!-\!1)$ blind-neighborhood.
Referring to the analysis of spatial correlation on real-world noise in AP-BSN~\cite{lee2022ap}, we set $k\!=\!5$ for 9$\times$9 blind-neighborhood.
More experimental analysis on the blind-neighborhood size can be seen in Sec.~\ref{sec:BNN-size}.
The details of network architecture are provided in the supplementary material.

\subsection{Supervision for Texture Areas} \label{sec:text_sup}

Note that we apply BNN to the whole image.
On the one hand, it is difficult to predetermine whether an area is flat or not from noisy images. 
Such an operation has little effect on the supervision prediction of flat areas.
On the other hand, textured areas can be easily detected from the full output of BNN for their supervision extraction.
In this subsection, we illustrate how to determine and extract supervision for textured areas.

\noindent\textbf{Determination of Textured Areas.} 
In flat areas, BNN sufficiently removes the noise, and the corresponding  variance is low.
In texture areas, although BNN is ambiguous in reserving details, the output is still not flat and the variance is high.
With the above observations, we can infer the flatness from the local variance of BNN output $\tilde{\textbf{x}}_1$.
Specifically, we densely extract image patches for each spatial location $(i, j)$, then calculate the standard deviation map $\bm{\sigma}$,
\begin{equation}
    \bm{\sigma}(i,j) = std(\tilde{\textbf{x}}_1\medmath{({i\!-\!\frac{n\!-\!1}{2}\!:\!i\!+\!\frac{n\!-\!1}{2},j\!-\!\frac{n\!-\!1}{2}\!:\!j\!+\!\frac{n\!-\!1}{2}}})),
\end{equation}
where $std(\cdot)$ represents the standard deviation function and is measured on 1-channel patches by averaging RGB values.
$n$ denotes the local window size, which we empirically set $n=7$.
Binarization $\bm{\sigma}(i, j)$ to determine the flatness (textured or flat) is rough and sometimes incorrect.
Instead, we convert the standard deviation map $\bm{\sigma}$ into soft coefficients $\bm{\alpha}$ which are normalized to $[0,1]$.
In our experiments, $\bm{\sigma}$ is a reliable indicator that it is usually high in the textured areas~(\eg, edges, texts) and is low in flat areas.
Io generate the coefficient $\bm{\alpha}$ that indicating flatness, we convert $\bm{\sigma}$ to $\bm{\alpha}$ with a piecewise function,
\begin{equation}
    \bm{\alpha}(i, j) = \begin{cases}
        S(\bm{\sigma}(i, j) - 1), & \bm{\sigma}(i, j)\leq l\\
        0.5, & l<\bm{\sigma}(i, j) \leq u\\
        S(\bm{\sigma}(i, j) - 5), & \bm{\sigma}(i, j)>u
    \end{cases}
\end{equation}
where $S(\cdot)$ denotes the Sigmoid function.
We empirically set $l=1$ and $u=5$ respectively.
Higher $\bm{\alpha}(i, j)$ means the local area is more textured.

\noindent\textbf{Locally Aware Network.}
Blind-neighborhood network performs poorly in texture areas, as it ignores the adjacent pixels which are essential for details prediction.
To extract proper supervision for texture areas, we present a locally aware network (LAN).
LAN is carefully designed to have a local receptive field, which focuses on recovering the texture details from neighboring pixels.
Nonetheless, LAN is a supervised network to be well-trained that requires clean supervision.
Fortunately, the flat areas of BNN output give a chance to supervise LAN.
On the one hand, the flat areas of BNN are well-denoised to approximate clean signal, and LAN can be trained safely.
On the other hand, The small receptive field of LAN makes it able to preserve texture details.
The loss function for training LAN is as follows:
\begin{equation}
    \mathcal{L}_{LAN} = (\bm{1 - \alpha})\cdot \left\| sg(\tilde{\textbf{x}}_1) - \tilde{\textbf{x}}_2 \right\|_1,
\end{equation}
where $\tilde{\textbf{x}}_1$ and $\tilde{\textbf{x}}_2$ is the output of BNN and LAN, respectively.
$sg(\cdot)$ denotes stop gradient operation~\cite{chen2021exploring}.
When training is done, LAN will have a transfer ability to predict the supervision for textured areas.

The network structure of LAN is simple yet delicate.
We stack 3$\times$3 convolution layers to create the local receptive field, specifically, $k$ 3$\times$3 layers can make up $(2k\!+\!1)\!\times\!(2k\!+\!1)$ receptive field.
In order to further refine the color information, we additionally add several 1$\times$1 convolution blocks with channel attention mechanism~\cite{zhang2018image}.
More details are provided in the supplementary material.

\subsection{Denoising with Learned Supervisions} \label{sec:denoising_network_training}
The learned images of BNN and LAN can be leveraged in two ways: image-level fusion or use as supervisions to train a denoising network.
We note that image-level fusion suffers from performance and efficiency issues.
From the performance perspective, although BNN and LAN are designed to denoise flat and texture areas respectively, they still have their weaknesses.
BNN may generate over-smooth results in flat areas, while LAN can not completely remove the noise in texture areas.
Image-level fusion can not avoid these drawbacks and only provides limited improvement.
Instead, utilizing the results as supervision to train a denoising network achieves a better trade-off between detail preserving and noise removal, which yields better performance.
From the efficiency perspective, BNN is computationally complex due to the special network design for blind-neighborhood.
Training additional denoising network provides more flexibility on the network complexity.
Thus, we tend to train a denoising network with learned spatially adaptive supervisions.
More experimental analysis about the two ways can be seen in Sec.~\ref{sec:image-level}.

We choose a common and representative network structure, \ie, U-Net~\cite{ronneberger2015u}, as our denoising network.
As shown in Figure~\ref{fig:overview}(c), the U-Net is trained with the following loss,
\begin{equation}
    \mathcal{L}_{D} = (\bm{1-\alpha})\cdot\left\| sg(\tilde{\textbf{x}}_1) - \hat{\textbf{x}} \right\|_1 + \bm{\alpha}\cdot\left\| sg(\tilde{\textbf{x}}_2) - \hat{\textbf{x}} \right\|_1,
    \label{equ:denoising}
\end{equation}
where $\hat{\textbf{x}}$ is the output of U-Net, $\tilde{\textbf{x}}_1$ and $\tilde{\textbf{x}}_2$ are the output of BNN and LAN, respectively.
$sg(\cdot)$ is stop gradient operation~\cite{chen2021exploring} and $\bm{\alpha}$ is the adaptive coefficients mentioned above.

\section{Experiments}
\subsection{Experimental Settings}
\textbf{Datasets.}
We conduct experiments on SIDD~\cite{abdelhamed2018sidd} and DND~\cite{plotz2017benchmarking}, which are widely used datasets for real-world image denoising.
SIDD dataset captures noisy-clean pairs with smartphone cameras.
Each noisy image is captured multiple times, and the mean image is served as the ground-truth.
It provides 320 image pairs (SIDD-Medium) for training, 1,280 patches for validation, and 1280 patches for benchmark testing.
DND~\cite{plotz2017benchmarking} dataset provides 50 image pairs for testing only.
Its noisy-clean pairs are conducted by shooting the same scene twice with different ISO settings.
The high-ISO images are taken as noisy inputs, while the corresponding low-ISO images are nearly noise-free and can serve as ground-truth.
The benchmark testing results (\ie, PSNR and SSIM) can be achieved by uploading the denoised patches to their official websites.
We note that SIDD and DND benchmarks provide evaluations in both the raw space and sRGB space.
As our method is developed to remove the spatially correlated noise of sRGB images, we do not compare with the self-supervised denoising methods~\cite{byun2021fbi, huang2021neighbor2neighbor, wang2022blind2unblind} for raw images.

\textbf{Implementation Details.}
We only leverage the noisy images of SIDD Medium dataset~\cite{abdelhamed2018sidd} to train our denoising framework.
Our BNN, LAN and the denoising network are trained successively with the same training settings.
We crop the training images into patches of size 256$\times$256, and augment the image patches with random flipping and rotation.
The augmented patches are formed as mini-batches of size 8 to facilitate network training.
Each network is trained with Adam optimizer~\cite{kingma2014adam} for 400k iterations, for a total of 1200k iterations.
The learning rate is initially set to $3\times 10^{-4}$, and decreased to zero with cosine annealing scheduler~\cite{loshchilov2016sgdr}.
%

\begin{table*}[t]
\small
\centering
\caption{Quantitative comparison of PSNR (dB), SSIM and LPIPS on SIDD~\cite{abdelhamed2018sidd} and DND~\cite{plotz2017benchmarking} datasets. LPIPS is calculated on SIDD validation dataset only as the ground-truth images on benchmark datasets are not available. Hereinafter, {\color{red}\textbf{red}} and {\color{blue}\underline{blue}} indicate the best and the second best results among unpaired and self-supervised methods, respectively.}
    \begin{tabular}{clccc}
    \toprule
         & \multirow{2}{*}{Method} & SIDD Validation & SIDD Benchmark & DND Benchmark\\
         & & PSNR$\uparrow$ / SSIM$\uparrow$ / LPIPS$\downarrow$ & PSNR$\uparrow$ / SSIM$\uparrow$ & PSNR$\uparrow$ / SSIM$\uparrow$\\
    \midrule
         \multirow{2}{*}{Non-learning based} & BM3D~\cite{dabov2007image} & 25.71 / 0.576 / 0.657 & 25.65 / 0.685 & 34.51 / 0.851\\
         & WNNM~\cite{gu2014weighted} & 26.05 / 0.592 / 0.635 & 25.78 / 0.809 & 34.67 / 0.865\\
    \midrule
         \multirow{3}{*}{\makecell[c]{Supervised\\(Synthetic pairs)}} & DnCNN~\cite{zhang2017beyond} & 26.21 / 0.604 / 0.712 & 26.25 / 0.599 & 32.43 / 0.790\\
         & CBDNet~\cite{guo2019toward} & 33.07 / 0.863 / 0.288 & 33.28 / 0.868 & 38.05 / 0.942\\
         & Zhou~\textit{et~al.}~\cite{zhou2020awgn} & 33.96 / 0.899 / 0.258 & 34.00 / 0.898 & 38.40 / 0.945\\
    \midrule
         \multirow{4}{*}{\makecell[c]{Supervised\\(Real pairs)}} & DnCNN~\cite{zhang2017beyond} & 37.73 / 0.943 / 0.245 & 37.61 / 0.941 & 38.73 / 0.945\\
         & Baseline, N2C\cite{ronneberger2015u} & 38.98 / 0.954 / 0.201 & 38.92 / 0.953 &  39.37 / 0.954\\
         & VDN~\cite{yue2019variational} & 39.29 / 0.956 / 0.208 & 39.26 / 0.955 & 39.38 / 0.952\\
         & Restormer~\cite{zamir2022restormer} & 39.93 / 0.960 / 0.198 & 40.02 / 0.960 & 40.03 / 0.956\\
    \midrule
         \multirow{4}{*}{Unpaired} & GCBD~\cite{chen2018image} & - & - & 35.58 / 0.922\\
         & UIDNet~\cite{hong2020end} & - & 32.48 / 0.897 & -\\
         & C2N~\cite{jang2021c2n} & 35.36 / {\color{blue}\underline{0.932}} / {\color{blue}\underline{0.192}} & 35.35 / {\color{red}\textbf{0.937}} & 37.28 / 0.924\\
         & Wu~\textit{et~al.}~\cite{wu2020unpaired} & - &  - & 37.93 / {\color{blue}\underline{0.937}}\\
    \midrule
        \multirow{7}{*}{Self-Supervised} & Noise2Void~\cite{krull2019noise2void} & 27.48 / 0.664 / 0.592 & 27.68 / 0.668 & -\\
        & Noise2Self~\cite{batson2019noise2self} & 29.94 / 0.782 / 0.556 & 29.56 / 0.808 & -\\
        & NAC~\cite{xu2020noisy} & - & - & 36.20 / 0.925\\
        & R2R~\cite{pang2021recorrupted} & - & 34.78 / 0.898 & -\\
        & CVF-SID~\cite{neshatavar2022cvf} & 34.15 / 0.911 / 0.423 & 34.71 / 0.917 & 36.50 / 0.924\\
        & AP-BSN+R$^3$~\cite{lee2022ap} & {\color{blue}\underline{36.74}} / {\color{red}\textbf{0.934}} / 0.281 & {\color{blue}\underline{36.91}} / 0.931 & {\color{blue}\underline{38.09}} / {\color{blue}\underline{0.937}}\\
        & Ours & {\color{red}\textbf{37.39}} / {\color{red}\textbf{0.934}} / {\color{red}\textbf{0.176}} & {\color{red}\textbf{37.41}} / {\color{blue}\underline{0.934}} & {\color{red}\textbf{38.18}} / {\color{red}\textbf{0.938}}\\
    \bottomrule
    \end{tabular}
    \label{tab:quantitative}
\end{table*}

\subsection{Results for Real-World Denoising}
\begin{figure*}
\newcommand{\siddsubfigurelen}{0.162}
\centering
    \begin{subfigure}[b]{\siddsubfigurelen\linewidth}
        \centering
        \includegraphics[width=\linewidth]{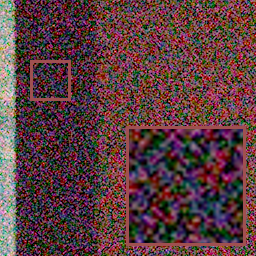}
    \end{subfigure}
    \hfill
    \begin{subfigure}[b]{\siddsubfigurelen\linewidth}
        \centering
        \includegraphics[width=\linewidth]{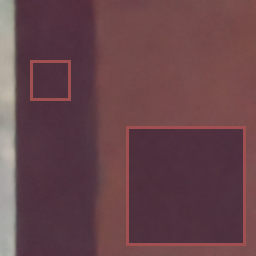}
    \end{subfigure}
    \hfill
    \begin{subfigure}[b]{\siddsubfigurelen\linewidth}
        \centering
        \includegraphics[width=\linewidth]{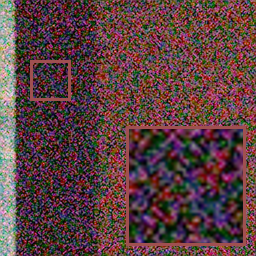}
    \end{subfigure}
    \hfill
    \begin{subfigure}[b]{\siddsubfigurelen\linewidth}
        \centering
        \includegraphics[width=\linewidth]{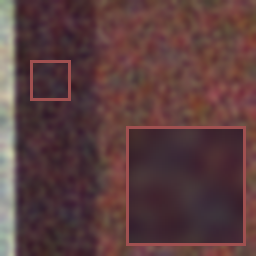}
    \end{subfigure}
    \hfill
    \begin{subfigure}[b]{\siddsubfigurelen\linewidth}
        \centering
        \includegraphics[width=\linewidth]{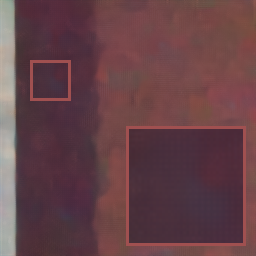}
    \end{subfigure}
    \hfill
    \begin{subfigure}[b]{\siddsubfigurelen\linewidth}
        \centering
        \includegraphics[width=\linewidth]{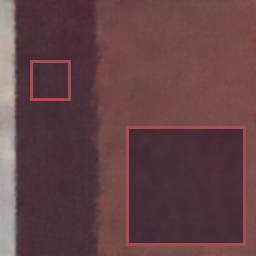}
    \end{subfigure}\\
    \begin{subfigure}[b]{\siddsubfigurelen\linewidth}
        \centering
        \includegraphics[width=\linewidth]{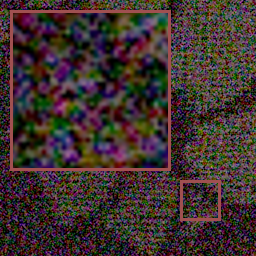}
        \caption{\centering\footnotesize Noisy Input}
    \end{subfigure}
    \hfill
    \begin{subfigure}[b]{\siddsubfigurelen\linewidth}
        \centering
        \includegraphics[width=\linewidth]{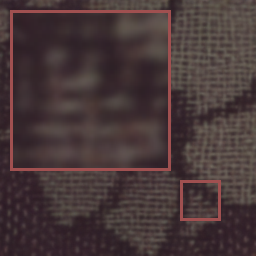}
        \caption{\centering\footnotesize Baseline, N2C~\cite{ronneberger2015u}}
    \end{subfigure}
    \hfill
    \begin{subfigure}[b]{\siddsubfigurelen\linewidth}
        \centering
        \includegraphics[width=\linewidth]{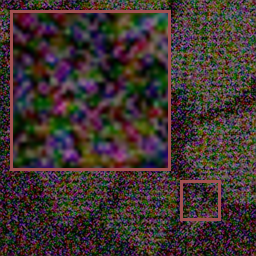}
        \caption{\centering\footnotesize Noise2Void~\cite{krull2019noise2void}}
    \end{subfigure}
    \hfill
    \begin{subfigure}[b]{\siddsubfigurelen\linewidth}
        \centering
        \includegraphics[width=\linewidth]{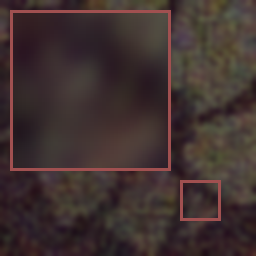}
        \caption{\centering\footnotesize CVF-SID~\cite{neshatavar2022cvf}}
    \end{subfigure}
    \hfill
    \begin{subfigure}[b]{\siddsubfigurelen\linewidth}
        \centering
        \includegraphics[width=\linewidth]{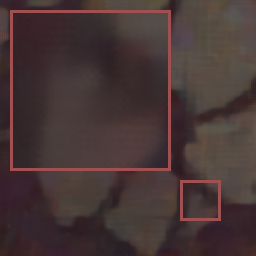}
        \caption{\centering\footnotesize AP-BSN+R$^3$~\cite{lee2022ap}}
    \end{subfigure}
    \hfill
    \begin{subfigure}[b]{\siddsubfigurelen\linewidth}
        \centering
        \includegraphics[width=\linewidth]{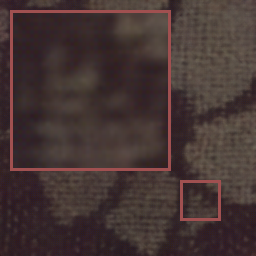}
        \caption{\centering\footnotesize Ours}
    \end{subfigure}
    \caption{Qualitative comparison on SIDD benchmark dataset~\cite{abdelhamed2018sidd}. Sub-figure (b) denotes the result of our denoising network trained on SIDD dataset in a supervised manner. Sub-figures (c)-(f) are from self-supervised denoising methods.}
    \label{fig:sidd}
\end{figure*}
\begin{figure*}
\newcommand{\dndsubfigurelen}{0.138}
    \begin{subfigure}[b]{0.276\linewidth}
        \centering
        \includegraphics[width=\linewidth]{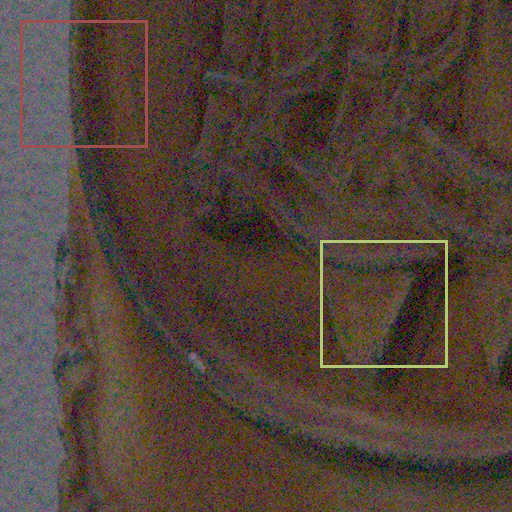}
        \caption{\centering\footnotesize Noisy Input}
    \end{subfigure}
    \hfill
    \begin{subfigure}[b]{\dndsubfigurelen\linewidth}
        \centering
        \includegraphics[width=\linewidth]{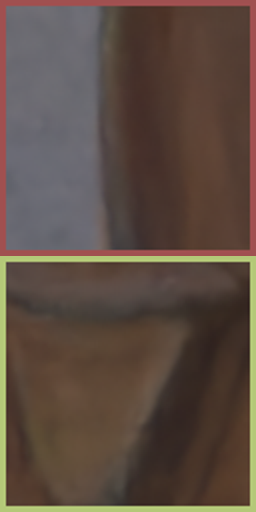}
        \caption{\centering\footnotesize {\scriptsize Baseline, N2C}~\cite{ronneberger2015u}}
    \end{subfigure}
    \hfill
    \begin{subfigure}[b]{\dndsubfigurelen\linewidth}
        \centering
        \includegraphics[width=\linewidth]{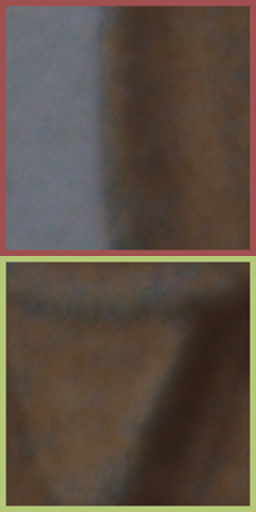}
        \caption{\centering\footnotesize NAC~\cite{xu2020noisy}}
    \end{subfigure}
    \hfill
    \begin{subfigure}[b]{\dndsubfigurelen\linewidth}
        \centering
        \includegraphics[width=\linewidth]{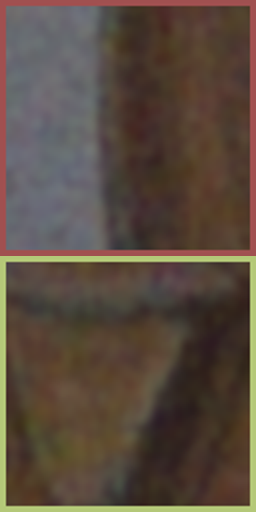}
        \caption{\centering\footnotesize CVF-SID~\cite{neshatavar2022cvf}}
    \end{subfigure}
    \hfill
    \begin{subfigure}[b]{\dndsubfigurelen\linewidth}
        \centering
        \includegraphics[width=\linewidth]{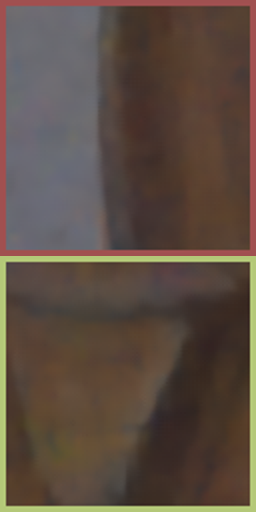}
        \caption{\centering\footnotesize AP-BSN+R$^3$~\cite{lee2022ap}}
    \end{subfigure}
    \hfill
    \begin{subfigure}[b]{\dndsubfigurelen\linewidth}
        \centering
        \includegraphics[width=\linewidth]{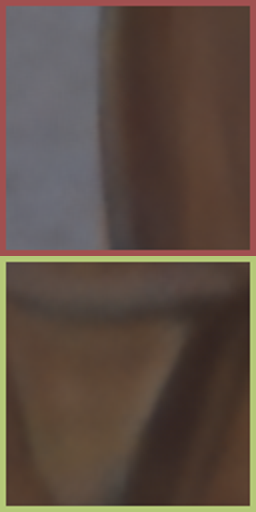}
        \caption{\centering\footnotesize Ours}
    \end{subfigure}
    \caption{\centering Qualitative comparison on DND benchmark dataset~\cite{plotz2017benchmarking}. Sub-figures (c)-(f) are from self-supervised denoising methods.}
    \label{fig:dnd}
\end{figure*}
\textbf{Quantitative Comparison.}
The quantitative comparison results of SIDD~\cite{abdelhamed2018sidd} and DND~\cite{plotz2017benchmarking} datasets can be seen in Table~\ref{tab:quantitative}.
Our method outperforms all unpaired and self-supervised approaches.
In terms of PSNR and SSIM metrics, our method is comparable to DnCNN~\cite{zhang2017beyond} trained with real-world pairs.
It demonstrates our method further mitigates the performance gap between self-supervised methods and supervised ones.
Among self-supervised methods, Noise2Void~\cite{krull2019noise2void} and Noise2Self~\cite{batson2019noise2self} fail to handle the noise in sRGB images due to their spatially independent noise assumption.
NAC~\cite{xu2020noisy} and R2R~\cite{pang2021recorrupted} depend on the parameters of the camera ISP to simulate training pairs, which are usually unavailable.
CVF-SID~\cite{neshatavar2022cvf}, AP-BSN~\cite{lee2022ap} and our method are directly trained on noisy images, while our method achieves the best performance.
We also provide LPIPS~\cite{zhang2018unreasonable} results on SIDD validation dataset to measure the perceptual quality.
Our method shows the lowest LPIPS score in all competing denoising methods.
It shows that our denoised images are most perceptually similar to the ground-truth.

\textbf{Qualitative Comparison.}
The visual comparison of state-of-the-art self-supervised methods on the benchmark datasets is shown in Figure~\ref{fig:sidd} and Figure~\ref{fig:dnd}.
As the clean ground-truth image is not provided, we train a supervised counterpart of our denoising network for reference, which is called baseline N2C.
From Figure~\ref{fig:sidd}, our method removes most spatially correlated noise and recovers the texture details better.
Other methods hardly generate visually pleasure results.
Noise2Void~\cite{krull2019noise2void} has little denoising effect and seems to converge into the noisy image.
CVF-SID~\cite{neshatavar2022cvf} can not decouple the noisy and clean components well from the noisy observations, leading to partially denoised results.
AP-BSN~\cite{lee2022ap} suffers from aliasing artifacts due to pixel-shuffle downsampling, \eg, the texture details on the fabric are missed.
The qualitative comparison on DND benchmark testing dataset~\cite{plotz2017benchmarking} is shown in Figure~\ref{fig:dnd}.
In comparison with other competitive methods, our result is cleaner and more photo-realistic.

\section{Ablation Study}
We conduct extensive ablation studies on SIDD validation dataset\cite{abdelhamed2018sidd} to analyze the effectiveness of the proposed framework.

\subsection{Effects of Supervision Components}
Table~\ref{tab:ablation_supervision} shows that each component of the proposed model is essential for the denoising network training.
From the table, $\tilde{\textbf{x}}_1$ and $\tilde{\textbf{x}}_2$ are not ideal supervisions and have their pros and cons.
Supervision by $\tilde{\textbf{x}}_1$ clearly removes the noise but at the risk of smoothing-out fine details.
Supervision by $\tilde{\textbf{x}}_2$ preserves the texture details but exhibits inferior denoising effect.
The network trained with both supervisions makes a
tradeoff between noise removal and texture preserving, thus
showing better results than separate ones.
Each one of them alone is not sufficient enough to facilitate the denoising network training.
Specifically, using BNN output as supervision alone leads to 0.46dB performance drop, and using LAN  output as supervision alone widens the gap to 1.35dB.
This demonstrates that both BNN and LAN provide indispensable supervisions for training the denoising network.
Adaptive coefficients $\bm{\alpha}$ is used for balancing the supervision strength of BNN and LAN outputs according to the local flatness.
We apply $\bm{\alpha}$ to the loss function (\ie, Eqn.~(\ref{equ:denoising})) to emphasize the BNN output in flat areas, as well as the LAN output in texture areas.
It provides 0.09dB performance boost, which demonstrates the effectiveness of learning spatially adaptive supervision for self-supervised denoising.

\subsection{Comparison with Image-Level Fusion}\label{sec:image-level}
\begin{figure}
    \centering
    \begin{subfigure}[b]{0.19\linewidth}
        \includegraphics[width=\linewidth]{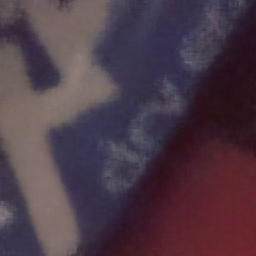}
        \caption{\centering $\tilde{\textbf{x}}_1$}
    \end{subfigure}
    \hfill
    \begin{subfigure}[b]{0.19\linewidth}
        \includegraphics[width=\linewidth]{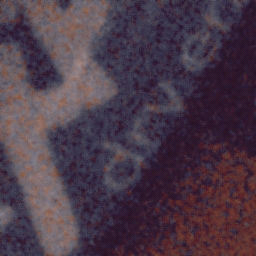}
        \caption{\centering $\tilde{\textbf{x}}_2$}
    \end{subfigure}
    \hfill
    \begin{subfigure}[b]{0.19\linewidth}
        \includegraphics[width=\linewidth]{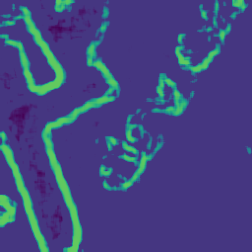}
        \caption{\centering $\bm{\alpha}$}
    \end{subfigure}
    \hfill
    \begin{subfigure}[b]{0.19\linewidth}
        \includegraphics[width=\linewidth]{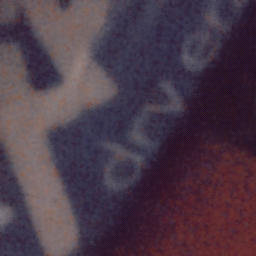}
        \caption{\centering Fused}
    \end{subfigure}
    \hfill
    \begin{subfigure}[b]{0.19\linewidth}
        \includegraphics[width=\linewidth]{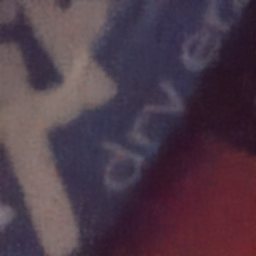}
        \caption{\centering $\hat{\textbf{x}}$}
    \end{subfigure}
    \caption{Visual comparison with image-level fusion strategy. (d) denotes the spatially adaptive fusion result $(\bm{1-\alpha})\cdot\tilde{\textbf{x}}_1 + \bm{\alpha}\cdot\tilde{\textbf{x}}_2$.}
    \label{fig:ablation_fusion}
\end{figure}

Apart from as supervisions, the outputs of BNN and LAN can also serve as denoised images directly, and further fused to get better results.
However, as illustrated in Sec.~\ref{sec:denoising_network_training}, the way of image-level fusion have performance and efficiency limits.
%
%
We try spatially adaptive fusion strategy that the outputs of BNN and LAN are weighted averaged according to the adaptive coefficients $\bm{\alpha}$, the result is shown in  Table~\ref{tab:ablation_image_fusion}.
Although the fusion strategy has improvement over individual images, it still falls 0.55dB from our proposed method.
Besides, due to the complexity of BNN, the inference time of the fusion strategy is much higher.
Figure~\ref{fig:ablation_fusion} provides visual comparison between the fused image and our denoising network output $\hat{\textbf{x}}$.
$\hat{\textbf{x}}$ shows better denoising effects.
It further demonstrates the effectiveness of taking BNN and LAN outputs as supervisions.

\begin{table}[t]
    \setlength{\tabcolsep}{2.3mm}
    \small
    \centering
    \caption{Ablation study of supervision components. The notation of $\tilde{\textbf{x}}_1$, $\tilde{\textbf{x}}_2$ and \bm{$\alpha$} follows Eqn.~(\ref{equ:denoising}).}
    \begin{tabular}{ccccc}
    \toprule
         Supervision of $\tilde{\textbf{x}}_1$ & $\checkmark$ & & $\checkmark$ & $\checkmark$\\
         Supervision of $\tilde{\textbf{x}}_2$  & & $\checkmark$ & $\checkmark$ & $\checkmark$\\
        Adaptive Coefficients \bm{$\alpha$} & & & & $\checkmark$\\
    \midrule
        PSNR of $\hat{\textbf{x}}$  & 36.84 & 35.95 & 37.30 & 37.39\\
    \bottomrule
    \end{tabular}
    \label{tab:ablation_supervision}
\end{table}
\begin{table}[t]
    \setlength{\tabcolsep}{2mm}
    \small
    \centering
    \caption{Comparison between image-level fusion strategy and our method.}
    \begin{tabular}{ccccc}
    \toprule
         Image & $\tilde{\textbf{x}}_1$ & $\tilde{\textbf{x}}_2$ & $(\bm{1-\alpha})\cdot\tilde{\textbf{x}}_1 + \bm{\alpha}\cdot\tilde{\textbf{x}}_2$ & $\hat{\textbf{x}}$\\
    \midrule
        PSNR & 36.37 & 35.00 & 36.84 & 37.39\\
        Time (ms) & 16.7 & 5.9 & 22.9 & 4.8\\
    \bottomrule
    \end{tabular}
    \label{tab:ablation_image_fusion}
\end{table}
\subsection{Blind-Neighborhood Size}\label{sec:BNN-size}
As illustrated in Sec.~\ref{sec:flat_sup}, the blind-neighborhood of BNN should cover the spatially correlated noisy pixels accurately.
Figure~\ref{fig:ablation_BNN_lan}(a) analyses the effects of blind-neightborhood size of BNN.
For 1$\times$1 blind-neighborhood size, BNN degrades to BSN which does not work for spatially correlated noise.
When gradually increasing the blind-neighborhood size, BNN shows better denoising effect as more noise-correlated pixels are excluded from the receptive field.
The max performance of BNN is achieved at 9$\times$9 blind-neighborhood size, which means all the noise correlation pixels have lain in this range.
It is consistent with the observation in AP-BSN~\cite{lee2022ap}.
Further increasing the blind-neighborhood size excludes noise-independent pixels from the receptive field, which is harmful to denoising performance.
The visualization of denoised images in Figure~\ref{fig:visualize_bnn_lan}(a) also supports the 9$\times$9 blind-neighborhood size.
In addition, the calculation of adaptive coefficients $\bm{\alpha}$ and the training of LAN are highly depend on the denoising quality of BNN on flat areas.
Thus, the performance of the denoising network shows the same trend as BNN.

\subsection{Local Receptive Size}
\begin{figure}[t]
    \centering
    \begin{subfigure}[b]{0.48\linewidth}
        \includegraphics[width=\linewidth]{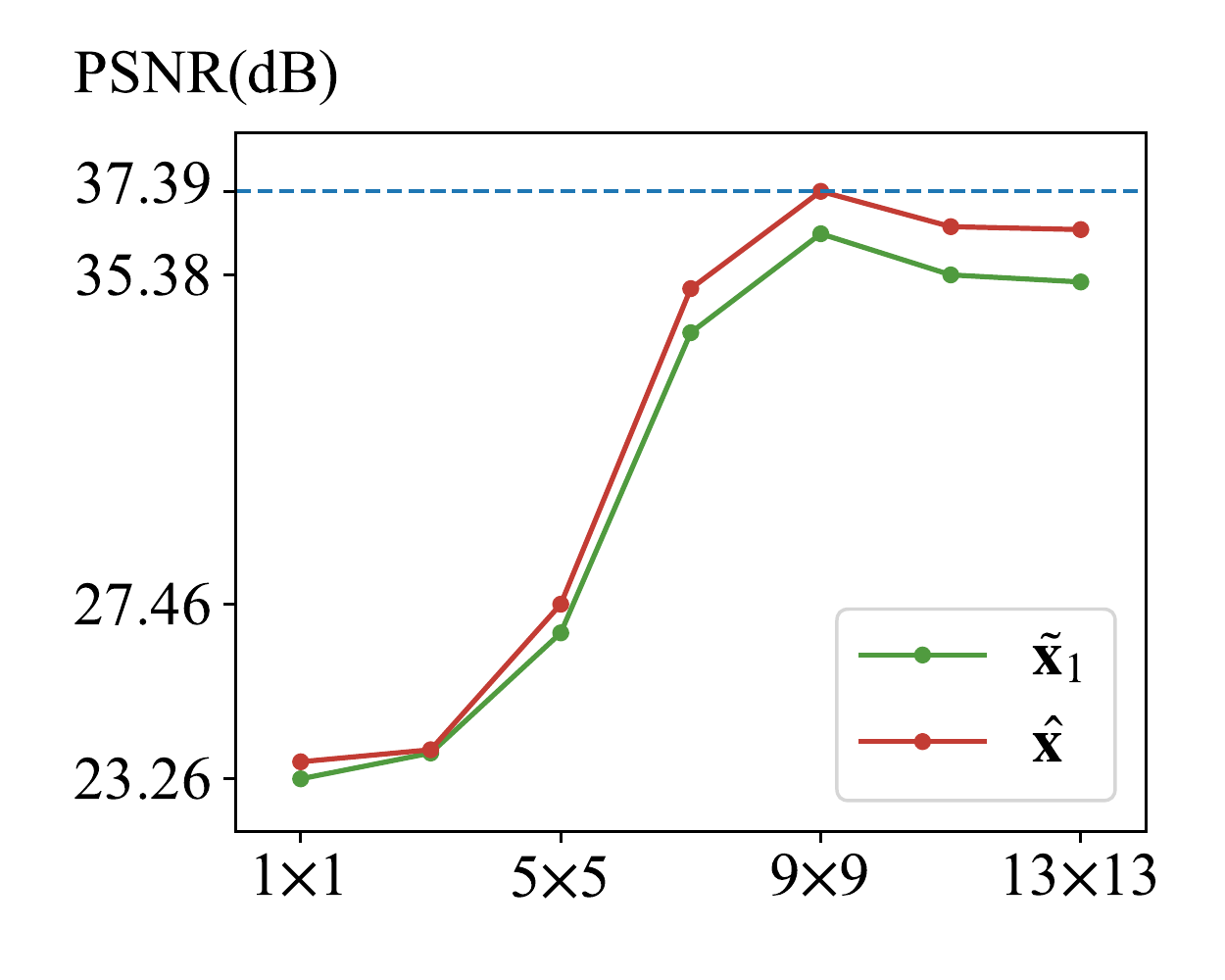}
        \caption{\centering Blind-neighborhood size.}
        \label{fig:ablation_bnn}
    \end{subfigure}
    \hfill
    \begin{subfigure}[b]{0.48\linewidth}
        \includegraphics[width=\linewidth]{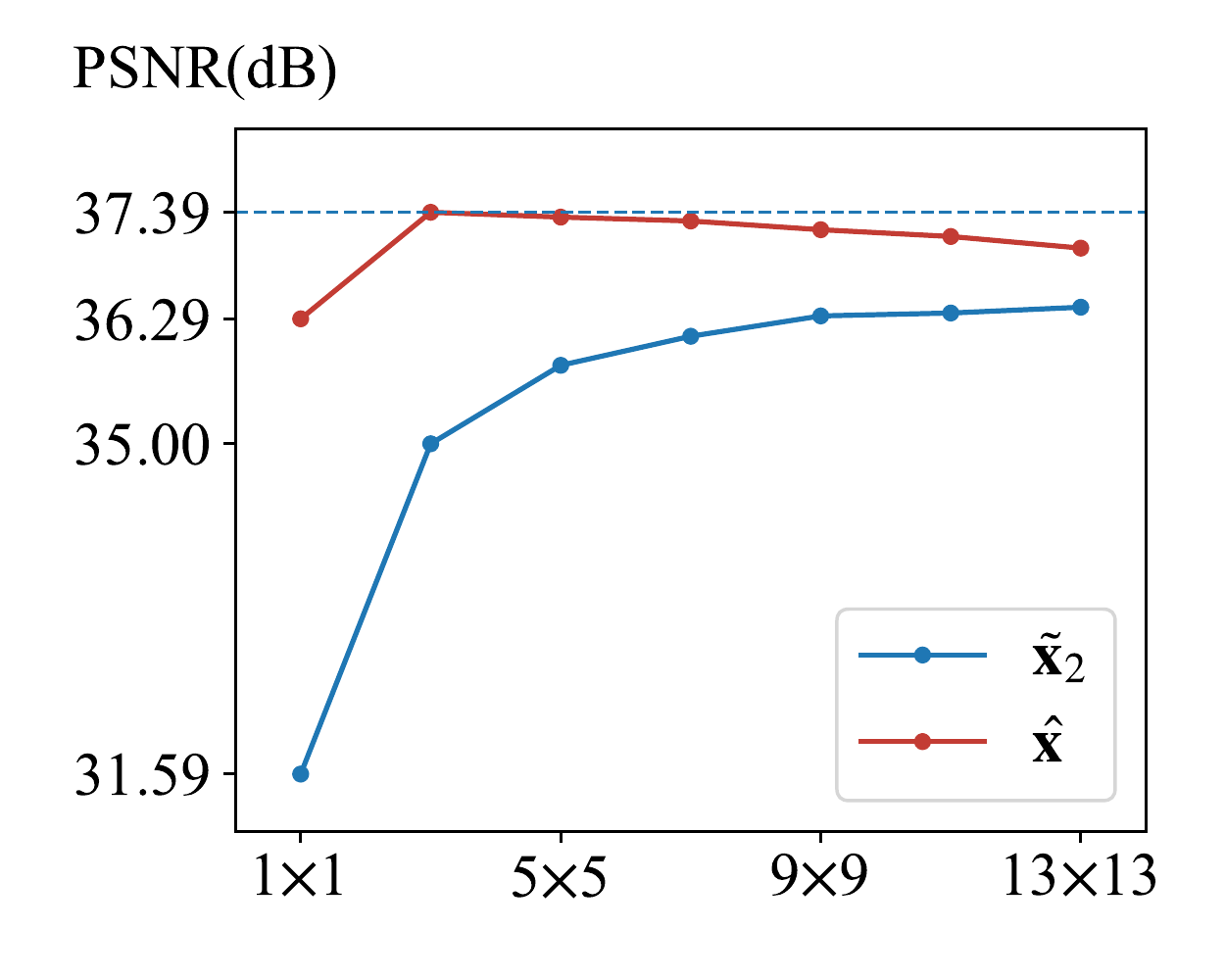}
        \caption{\centering Local receptive size.}
        \label{fig:ablation_lan}
    \end{subfigure}
    \caption{Effects of the blind-neighborhood size and local receptive size. (a) BNN achieves best performance with blind-neighborhood size 9$\times$9. (b) LAN achieves better performance as the local receptive size increases, but the final denoising network performs best with local receptive size 3$\times$3.}
    \label{fig:ablation_BNN_lan}
\end{figure}

\begin{figure}
\newcommand{\ablationsubfigurelen}{0.15}
    \centering
    \begin{subfigure}[b]{\ablationsubfigurelen\linewidth}
        \caption*{\centering 1$\times$1}
        \includegraphics[width=\linewidth]{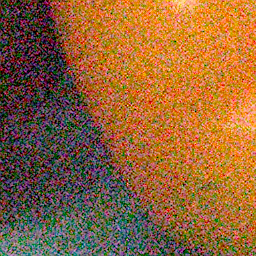}
    \end{subfigure}
    \hfill
    \begin{subfigure}[b]{\ablationsubfigurelen\linewidth}
        \caption*{\centering 3$\times$3}
        \includegraphics[width=\linewidth]{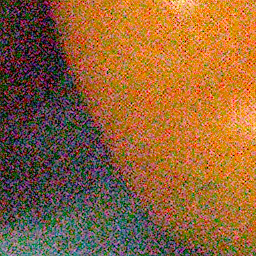}
    \end{subfigure}
    \hfill
    \begin{subfigure}[b]{\ablationsubfigurelen\linewidth}
        \caption*{\centering 5$\times$5}
        \includegraphics[width=\linewidth]{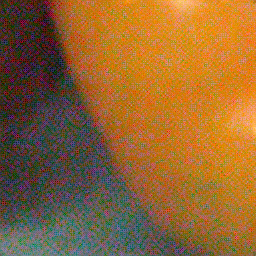}
    \end{subfigure}
    \hfill
    \begin{subfigure}[b]{\ablationsubfigurelen\linewidth}
        \caption*{\centering 7$\times$7}
        \includegraphics[width=\linewidth]{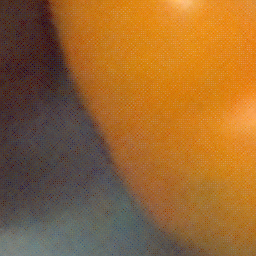}
    \end{subfigure}
    \hfill
    \begin{subfigure}[b]{\ablationsubfigurelen\linewidth}
        \caption*{\centering 9$\times$9}
        \includegraphics[width=\linewidth]{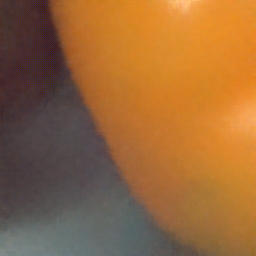}
    \end{subfigure}
    \hfill
    \begin{subfigure}[b]{\ablationsubfigurelen\linewidth}
        \caption*{\centering 11$\times$11}
        \includegraphics[width=\linewidth]{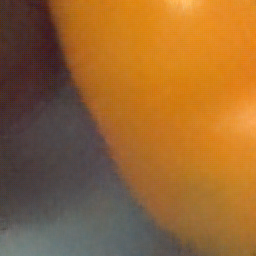}
    \end{subfigure}\\
    \centering {\footnotesize (a) BNN output with different blind-neighborhood sizes.}\\
    \begin{subfigure}[b]{\ablationsubfigurelen\linewidth}
        \includegraphics[width=\linewidth]{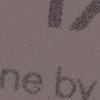}
    \end{subfigure}
    \hfill
    \begin{subfigure}[b]{\ablationsubfigurelen\linewidth}
        \includegraphics[width=\linewidth]{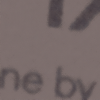}
    \end{subfigure}
    \hfill
    \begin{subfigure}[b]{\ablationsubfigurelen\linewidth}
        \includegraphics[width=\linewidth]{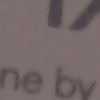}
    \end{subfigure}
    \hfill
    \begin{subfigure}[b]{\ablationsubfigurelen\linewidth}
        \includegraphics[width=\linewidth]{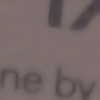}
    \end{subfigure}
    \hfill
    \begin{subfigure}[b]{\ablationsubfigurelen\linewidth}
        \includegraphics[width=\linewidth]{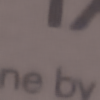}
    \end{subfigure}
    \hfill
    \begin{subfigure}[b]{\ablationsubfigurelen\linewidth}
        \includegraphics[width=\linewidth]{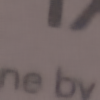}
    \end{subfigure}\\
    \centering {\footnotesize (b) LAN output with different local receptive sizes.}
    \caption{Visual comparison of different blind-neighborhood sizes and local receptive sizes.}
    \label{fig:visualize_bnn_lan}
\end{figure}

Figure~\ref{fig:ablation_BNN_lan}(b) and \ref{fig:visualize_bnn_lan}(b) show the effects of local receptive size of LAN.
For 1$\times$1 local receptive size, LAN has little denoising effect due to no neighboring pixels are leveraged.
For larger local receptive sizes, LAN can effectively predict the favorable signal from neighbor pixels with good details.
But at the same time, the output of LAN tends to be more blurry.
The outputs of LAN should serve as good supervision of texture areas for the denoising network training.
The best performance of our denoising framework is achieved when LAN has 3$\times$3 local receptive size.

\section{Conclusion}
In this work, we propose a novel perspective to solve real-world image self-supervised denoising, \ie, seeking for spatially adaptive supervision for a denoising network according to the image characteristics. 
The clean signal in flat areas can be inferred from non-neighboring noise-independent pixels, while the texture details  should come from neighboring ones.
Thus, blind-neighborhood network (BNN) is proposed to learn the supervision for flat areas.
Adaptive coefficients and locally aware network (LAN) are proposed to determine the flatness and learn the supervision for texture areas, respectively.
The denoising network trained by learned supervisions outperforms state-of-the-art self-supervised and unpaired methods with better efficiency.

~\\
\noindent\textbf{Acknowledgement}.
This work is partially supported by the National Natural Science Foundation of China(NSFC) under Grant No. U19A2073.

\clearpage
{\small
\bibliographystyle{ieee_fullname}
\bibliography{egbib}
}

\end{document}